%% file: lrec2022.tex
\documentclass[10pt, a4paper]{article}
\usepackage{lrec2022} % this is the new LREC2022 Style
\usepackage{multibib}
\newcites{languageresource}{Language Resources}
\usepackage{graphicx}
\usepackage{tabularx}
\usepackage{soul}
\usepackage{titlesec}
\titleformat{\section}{\normalfont\large\bfseries\center}{\thesection.}{1em}{}
\titleformat{\subsection}{\normalfont\SmallTitleFont\bfseries\raggedright}{\thesubsection.}{1em}{}
\titleformat{\subsubsection}{\normalfont\normalsize\bfseries\raggedright}{\thesubsubsection.}{1em}{}
\renewcommand\thesection{\arabic{section}}
\renewcommand\thesubsection{\thesection.\arabic{subsection}}
\renewcommand\thesubsubsection{\thesubsection.\arabic{subsubsection}}
\usepackage{epstopdf}
\usepackage[utf8]{inputenc}

\usepackage{hyperref}
\usepackage{xstring}
\usepackage{amsmath}
\usepackage{makecell}
\usepackage{CJKutf8}
\usepackage{color}
\usepackage{booktabs}

\title{MMChat: Multi-Modal Chat Dataset on Social Media}

\name{
  Yinhe Zheng\textsuperscript{$\spadesuit$}\sthanks{\hspace{0.2cm}Equal Contribution}, 
  Guanyi Chen\textsuperscript{$\clubsuit$}\footnotemark[1], 
  Xin Liu\textsuperscript{$\heartsuit$}, 
  Jian Sun\textsuperscript{$\spadesuit$}\sthanks{\hspace{0.2cm}Corresponding Author}
} 

\address{\textsuperscript{$\spadesuit$} Alibaba Group, \textsuperscript{$\clubsuit$} Utrecht University, \textsuperscript{$\heartsuit$}Samsung Research China - Beijing (SRC-B)\\
\texttt{zhengyinhe1@163.com, g.chen@uu.nl, jian.sun@alibaba-inc.com}}

\abstract{
Incorporating multi-modal contexts in conversation is important for developing more engaging dialogue systems. In this work, we explore this direction by introducing \textsc{MMChat}: a large-scale Chinese multi-modal dialogue corpus (32.4M raw dialogues and 120.84K filtered dialogues). Unlike previous corpora that are crowd-sourced or collected from fictitious movies, \textsc{MMChat} contains image-grounded dialogues collected from real conversations on social media, in which the \emph{sparsity} issue is observed. Specifically, image-initiated dialogues in common communications may deviate to some non-image-grounded topics as the conversation proceeds. To better investigate this issue, we manually annotate 100K dialogues from \textsc{MMChat} and further filter the corpus accordingly, which yields \textsc{MMChat-hf}. We develop a benchmark model to address the sparsity issue in dialogue generation tasks by adapting the attention routing mechanism on image features. Experiments demonstrate the usefulness of incorporating image features and the effectiveness of handling the sparsity of image features.
 \\ \newline  }

\begin{document}

\maketitleabstract

\input{sec/intro}
\input{sec/dataset}
\input{sec/model}
\input{sec/experiment}

\section{Conclusion}

We introduce \textsc{MMChat}, a large-scale multi-modal dialogue corpus that reveals the image-sparsity phenomenon in real conversations. Our dataset contains 120.84K dialogue sessions filtered from 32.4M sessions of raw multi-modal dialogues. Building on 100K dialogues from \textsc{MMChat}, we further conduct human filtering, yielding \textsc{MMChat-hf}, in which there are 19.9K high-quality multi-modal dialogue sessions. A dialogue model is proposed to tackle the image-sparsity issue utilizing \textsc{MMChat}. Experiment results indicate that both \textsc{MMChat} and \textsc{MMChat-hf} help to develop image-grounded dialogue systems and facilitate further study of the image-sparsity issue. Besides the filtered dialogues in \textsc{MMChat}, we will also release all the raw dialogues obtained in the data collection process to facilitate further studies.

\section{Broader Considerations}

Our dataset \textsc{MMChat} originates from a Chinese social media. The dataset collection and release protocols are carefully designed to avoid violating the privacy of each user on that social media. Specifically, each user's permission setting is strictly respected so that only publicly visible contents are collected. Rules are designed to filter out dialogues that may potentially expose users' private information, such as phone numbers or emails. Moreover, we will not host these images in \textsc{MMChat} on our own server. Only the URLs to these images will be released along with the download scripts.

To further enforce the data privacy, \textsc{MMChat} is released under strict terms for academic uses only, in which they promise no abuse of \textsc{MMChat} besides academic purposes.

In addition to the privacy issues, there might also be toxic or biased texts in \textsc{MMChat} or be generated by MMDSs trained on \textsc{MMChat}. Although we take the responsibility to remove toxic texts (using an offensive word list, an offensive content classifier, and human filtering), we cannot guarantee that there are no offensive contents left. However, as offensive and abusive content recognition is a rapidly developing area~\cite{vidgen2019challenges}, we would deploy more advanced filters once the new state-of-the-art offensive and abusive classifiers are proposed in the future.

Regarding the potential biases, except those from the dataset itself~\cite{henderson2018ethical} (which always exists in dialogue datasets), biases might be introduced by the pre-trained language model~\cite{bender2021dangers} and the pre-trained image encoder~\cite{steed2020image} used in this work. In the future, we head to apply and develop corresponding mitigation techniques~(following works such as~\newcite{dinan2019queens} and~\newcite{liu2020mitigating}).

During annotation, we pay each annotator 0.5 CNY per item. This results in approximately 60 CNY per hour, which is 4.5 times the minimum wage standard in China.

Besides, we also note that the goal of our work is to facilitate further work on multi-modal dialogue systems. Although the model used in this work is still far from realistic, our dataset can be regarded as an initial step toward the sparsity issue in real-world multi-modal conversations.

\section{Bibliographical References}\label{reference}

\bibliographystyle{lrec2022-bib}
\bibliography{acl2021,anthology}

%\section{Language Resource References}
%\label{lr:ref}
%\bibliographystylelanguageresource{lrec2022-bib}
%\bibliographylanguageresource{languageresource}

\end{document}

%% file: sec/intro.tex
\section{Introduction}

The ability to converse like a human is one of the desiderata for building open-domain dialogue systems~\cite{adiwardana2020towards}. Current attempts to build human-like open-domain dialogue systems generally follow two angles: 1) Enriching the dialogue system with textual or structural contexts such as knowledge~\cite{madotto-etal-2018-mem2seq} or personalities~\cite{zhang-etal-2018-personalizing,zheng2019personalized}; 2) Enabling the dialogue systems to perceive multi-modality contexts beyond text such as vision, voice, or even gesture~\cite{shuster2019dialogue,shuster2020multi,liao2018knowledge,ju2019all}. Systems built following the second angle are also known as Multi-Modal Dialogue Systems (MMDSs).

To facilitate the development of data-driven MMDSs, a few dialogue datasets containing visual information have been constructed \cite{mostafazadeh-etal-2017-image,mogadala2019trends,alamri2019audio,kottur2019clevr,pasunuru2018game,shuster-etal-2020-image,meng2020openvidial}. For instance, \newcite{shuster-etal-2020-image} introduced a crowd-sourced image grounded dialogue corpus \textsc{Image-Chat}, in which annotators are employed to chat in accordance with given images. \newcite{meng2020openvidial} proposed \textsc{OpenViDial} by directly extracting dialogues and their visual contexts from movies and TV series. There are also works on visual question answering~\cite{das2017visual} that focus on the question answering tasks involving image inputs.

\begin{figure}[t]
  \centering
  \includegraphics[width=210px]{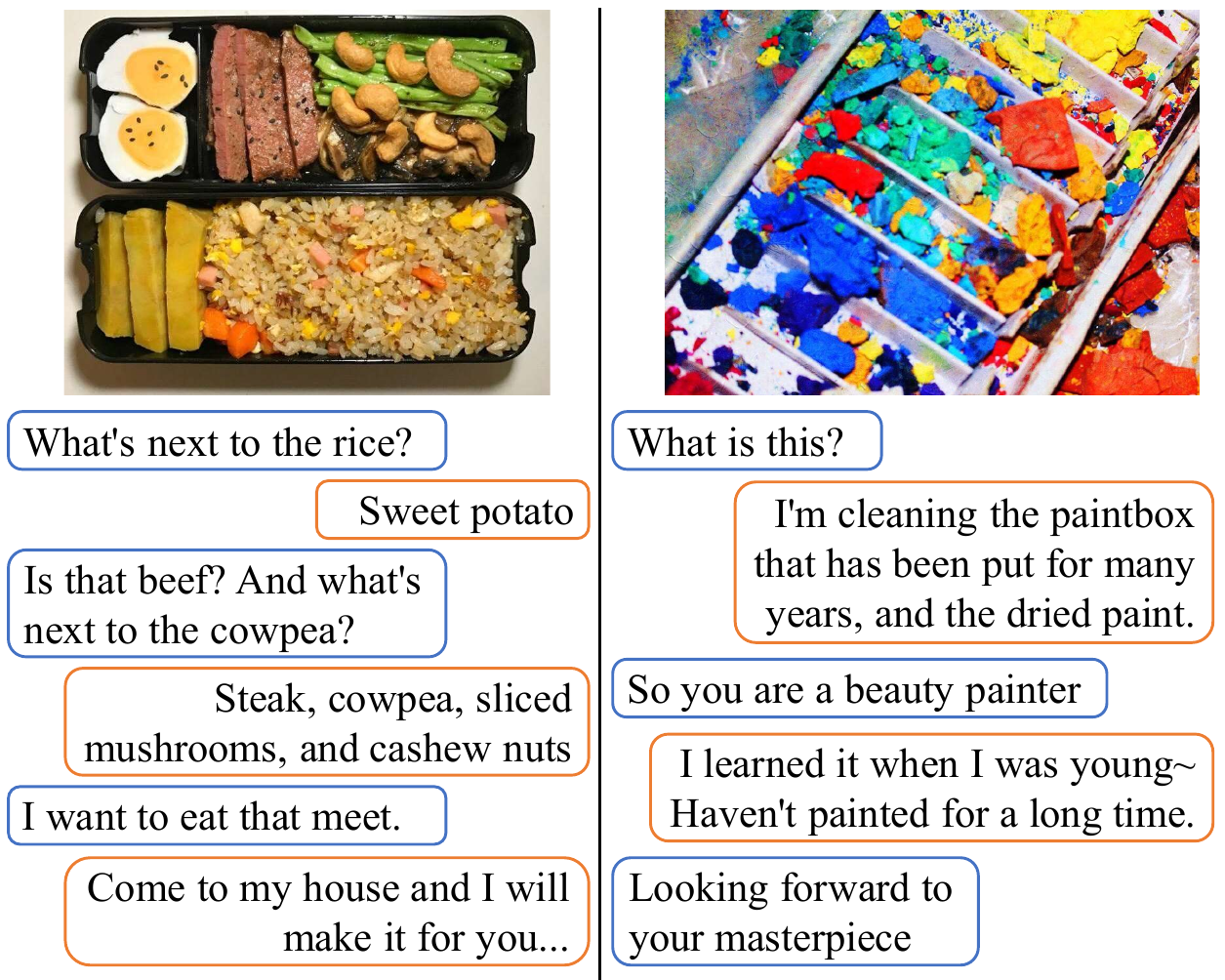}
  \caption{Example dialogues from \textsc{MMChat} (translated from Chinese).}
  \label{fig:image_dialog_example}
\end{figure}

A significant drawback of existing datasets is that they postulated every utterance in a dialogue to be grounded on the given image. Nevertheless, this is not always true in our daily communications. Concretely, the topic triggered by the image may drift in the conversation flow so that not every utterance in a dialogue session is image-grounded. Taking the right dialogue in Figure~\ref{fig:image_dialog_example} as an example, it is initialized by objects shown in the associated image (i.e., ``\emph{paintbox}'' and ``\emph{dried paint}'') but the focus of the following dialogue move to the speaker's own experience as a painter, which is not image-related anymore. A similar pattern is also observed in the left example of Figure~\ref{fig:image_dialog_example}. We refer to this phenomenon as the issue of \emph{sparsity} and dialogues that exhibit this phenomenon as the \emph{sparse image-grounded dialogues}. 

To tackle the above issue, we introduce \textsc{MMChat}: a large-scale dataset that contains sparse image-grounded dialogues in Chinese. We first collected 32.4M sessions of raw dialogues and 8.41M associated images from social media. Based on these raw dialogues, we design an elaborate data filtering process and construct \textsc{MMChat}, which contains 120.84K sessions of filtered high-quality dialogues and 204.32K images. Dialogues that are incoherent or involve offensive content are filtered. Two example dialogue sessions are shown in Figure~\ref{fig:image_dialog_example}. Unlike previous multi-modal dialogue datasets that only provide a single raw image per dialogue session, each filtered dialogue session in \textsc{MMChat} corresponds to one or multiple images. The semantic information of each image is further revealed in our dataset using a pre-trained image caption model. Specifically, a set of detected object labels and a generated descriptive caption are released for each image in \textsc{MMChat}.

To further improve the quality of dialogues in \textsc{MMChat}, we sample 100K dialogue sessions from \textsc{MMChat}, and manually check the quality of images and whether these dialogues are strongly correlated with the associated images. This yields a ``human filtered'' dataset \textsc{MMChat-hf} that contains 19.90K dialogue sessions and 52.24K images.

Building on both \textsc{MMChat} and \textsc{MMChat-hf}, We provide a strong benchmark model to tackle the image-sparsity issue in open-domain dialogue generation tasks based on the attention routing mechanism \cite{zheng2020pre}. Evaluation results on both datasets suggest that incorporating visual contexts contributes positively to dialogue modeling, and the approach used in our benchmark model helps alleviate the sparsity issue.

Besides enlightening advanced models for realistic multi-modal conversations, \textsc{MMChat} is also built to help understand how Chinese multi-modal communications are conducted from the aspect of social science~\cite{jovanovic2018multimodal}. The vast amount of dialogues and images in \textsc{MMChat} can also benefit the study of multi-modal pretraining models. 

In what follows, we summarize our main contributions:
\begin{itemize}
    \item We construct a large multi-modal dialogue dataset \textsc{MMChat}, addressing the issue of ``sparsity''. A dedicated automatic filtering process is proposed to clean the dataset.
    \item We offer a human filtered dataset \textsc{MMChat-hf} based on 100K dialogue sessions sampled from \textsc{MMChat}.
    \item We build benchmark models on \textsc{MMChat}. The results indicate that incorporating visual contexts contribute positively to dialogue modeling, and our benchmark model can better tackle the sparsity issue.
\end{itemize}

Our dataset and code are available in \url{https://github.com/silverriver/MMChat}

%% file: sec/dataset.tex
\section{Dataset Construction: \textsc{MMChat}} \label{sec:dataset}

\textsc{MMChat} originates from a Chinese social media on which users can share their daily lives through images and texts. This section starts by introducing how \textsc{MMChat} is constructed (Section~\ref{sec:collect}) and cleaned (Section~\ref{sec:clean}). We then sketch how to manually annotate \textsc{MMChat} to produce \textsc{MMChat-hf} (Section~\ref{sec:annotation}). At length, we report an analysis of the \textsc{MMChat} dataset.

\subsection{Data Collection} \label{sec:collect}

A two-phase pipeline is used to construct raw dialogues in \textsc{MMChat}: the first phase aims to collect seed users who are active on social media. We start this phase with a few hand-collected mass media accounts. Professionals maintain these accounts and are committed to posting daily news on broad topics. The users who comment under this news are collected as our \emph{seed users}. The second phase starts from the seed users collected above. Specifically, the images posted by these seed users are obtained, and the comments under these images are collected. Dialogues along these images are constructed by restoring the reply relationship between these comments.

The two-phase data collection approach used in our study effectively avoids spammers' noises since most spammers will not bother to follow and reply to daily news. Moreover, we also filter out seed users that are not active to make the data collection process more effective. Finally, we collected a corpus containing about 32.4M sessions of raw dialogues.

\subsection{Data Filtering and Post-processing} \label{sec:clean}

\begin{table}[t]
  \centering
  \setlength\tabcolsep{2.0pt} % default value: 6pt
  \begin{tabular}{lr}
  \toprule
    \#(Dialogues)           & 120.84K \\
    \#(Total Images)              & 204.32K \\  
    \#(Total Utterances)          & 314.13K \\
    \#(Dialogue Sessions) Longer than 4 & 17.32K \\
    \#(Image) per Dialogue        & 2.91    \\
    \#(Utterance) per Dialogue    & 2.59    \\
    \#(Character) per Utterance   & 8.52    \\
    \midrule
    \#(Raw Dialogues) & 32.4M   \\
  \bottomrule
  \end{tabular}
  \caption{Statistics of \textsc{MMChat}.}
  \label{tab:data_info}
\end{table}

\begin{CJK}{UTF8}{gbsn}
\begin{table*}[t]
    \centering
    \small
    \begin{tabular}{clcl}
    \toprule
    %Image with bounding boxes& Meta-Information \\ \midrule
    \begin{minipage}{.2\textwidth} % 1656
      \includegraphics[scale=0.4]{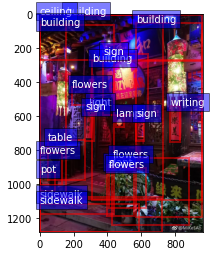}
     % [' 建筑', ' 灯', ' 人行道、人行道', ' 天花板', '红色的 签名', ' 签名', '红色的 写作', ' 植物', ' 花', '黄色的 ', ' 光', '蓝色的 ', '红色的 ', ' 桌子', ' 锅']
    \end{minipage} & \makecell[l]{\textbf{Caption}: A store on the \\street during night. \\ \textbf{Detected Objects}: building, \\light, red signature, \\ red calligraphy, plants, \\flowers, lamps, table}  &
        \begin{minipage}{.2\textwidth}
      \includegraphics[scale=0.4]{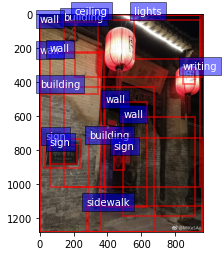}
     % [' 建筑', ' 墙', '黑色的 签名', ' 天花板', ' 签名', '红色的 签名', ' 写作', ' 灯', ' 人行道、人行道']
    \end{minipage} & \makecell[l]{\textbf{Caption}: A brick house \\with guidepost. \\ \textbf{Detected Objects}: building, \\wall, black signature, \\ red signature, calligraphy, lamp} \\ & \\
    \begin{minipage}{.22\textwidth}
    \includegraphics[scale=0.3]{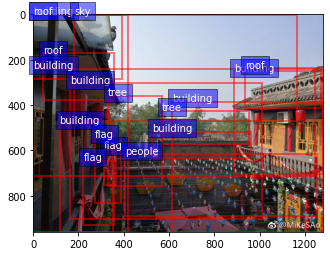}
     % [' 建筑', '蓝色的 天空', '红色的 旗帜', ' 屋顶', ' 人', ' 旗帜', ' 树']
    \end{minipage} & \makecell[l]{\textbf{Caption}: Flags on the street. \\ \textbf{Detected Objects}: building, \\blue sky, red flags, roof, \\ people, trees} \\ & \\
    % ['去王家大院看看', '跟古城差不多吧', '比古城好，有时间可以去', '好的~古城过度开发了都成商业街了', '嗯嗯，王家大院很震撼我']
    \multicolumn{4}{l}{\makecell[l]{A: 去王家大院看看。\\ \quad\ (Will you have a visit to the Wang Family Courtyard?) \\ B: 跟古城差不多吧。\\ \quad\ (Isn't there having the same scene with the ancient city?) \\ A: 比古城好，有时间可以去。\\ \quad\ (Much better than the ancient city. You can have a visit if you have spare time.) \\ B: 好的~古城过度开发了都成商业街了。\\ \quad\ (Sure. The ancient city has been over-commercialized.) \\ A: 嗯嗯，王家大院很震撼我。\\ \quad\ (Uh-huh. The Wang Family Courtyard impressed me a lot.)}} \\
    \bottomrule
    \end{tabular}
    \caption{An example dialogue (translated from Chinese) and its associated images from \textsc{MMChat}. Each image's semantic information (including object labels, attributes, bounding boxes and image captions) is provided.}
    \label{tab:ex2}
\end{table*}
\end{CJK}

To improve the quality of \textsc{MMChat}, a set of rules is carefully designed to filter out low-quality images and dialogues from the raw corpus collected in Section~\ref{sec:collect}. Specifically, images with extremely low resolution (fewer than $500$ pixels) or high aspect ratios (larger than $10$) are abandoned, and dialogues that contain extremely long utterances (longer than 200 tokens) are filtered. Moreover, we only retain dialogues that contain more than $3$ utterances. The offensive contents are also filtered using an offensive word list and a pre-trained offensive content classifier~\cite{wang2020chinese}.

To ensure the dialogue contents in \textsc{MMChat} are related to the corresponding images in the first few turns of the conversation, we only retain images that are uploaded through the direct-share mode. This mode allows users to share images without providing textual content. We argue that the initial few turns of the dialogues following these image-only posts are usually triggered by the visual information because there are no previous textual contexts except for the uploaded images.

Note that eliminating posts that are not uploaded through the direct-share mode filters out a vast majority of collected raw dialogues. However, this rule is adopted not because these filtered dialogues are of low quality but because we only have limited computation resources. We want our model to focus on dialogues that are more closely related to its multi-modal contexts. We believe these filtered raw dialogues are useful in building large-scale multi-modal dialogue models or multi-modal pre-trained models. We will release all the collected raw dialogues to facilitate further studies in this direction.

The statistics of the resulting \textsc{MMChat} dataset are shown in Table \ref{tab:data_info}. Each dialogue session is associated with at least one image (9 images maximum), and a considerable amount of sessions (more than 17.32K) in \textsc{MMChat} contain at least 4 utterances (i.e., 2 turns). Note that different dialogues may share the same post (i.e., the same set of images). To protect data privacy, \textsc{MMChat} is released under strict terms for academic users only. More details for the data release protocol can be found in the Broader Consideration section.

\subsection{Human Filtering} \label{sec:annotation}

\begin{table}[t]
  \centering
  \setlength\tabcolsep{2.0pt} % default value: 6pt
  \begin{tabular}{lr}
  \toprule
    \#(Dialogues)           & 19.90K \\
    \#(Total Images)              & 52.66K \\  
    \#(Total Utterances)          & 81.06K \\
    \#(Dialogue Sessions) Longer than 4 & 8.91K \\
    \#(Image) per Dialogue        & 2.70    \\
    \#(Utterance) per Dialogue    & 4.07    \\
    \#(Character) per Utterance   & 11.93    \\
  \bottomrule
  \end{tabular}
  \caption{Statistics of \textsc{MMChat-hf}.}
  \label{tab:mmachat_hf_info}
\end{table}

To facilitate further studies on \textsc{MMChat}, we recruit annotators to manually filter \textsc{MMChat}, and construct a dataset \textsc{MMChat-hf} with higher quality. Concretely, we random sample 100k dialogue sessions from \textsc{MMChat} and ask annotators to annotate each session from the following three aspects: 1) \emph{Whether the associated images are qualified}. The associated images of a dialogue session are not qualified if any of the images is overlong/flat (i.e., depth-width ratio $>$ 10 or $<$ 0.1) or is a screenshot of texts (e.g., email, news, etc.). We also identify selfies and offensive images as disqualified; 2) \emph{Whether the dialogue contents are non-offensive}. Though we have filtered offensive contents automatically, we ask annotators to check them further manually; 3) \emph{Whether the dialogue content is strongly correlated with the associated images}. A dialogue session is annotated as ``true'' in this aspect if its content contains mentions of any object/person/background of its associated images.

\begin{figure*}[t]
    \centering
    \includegraphics[width=400px]{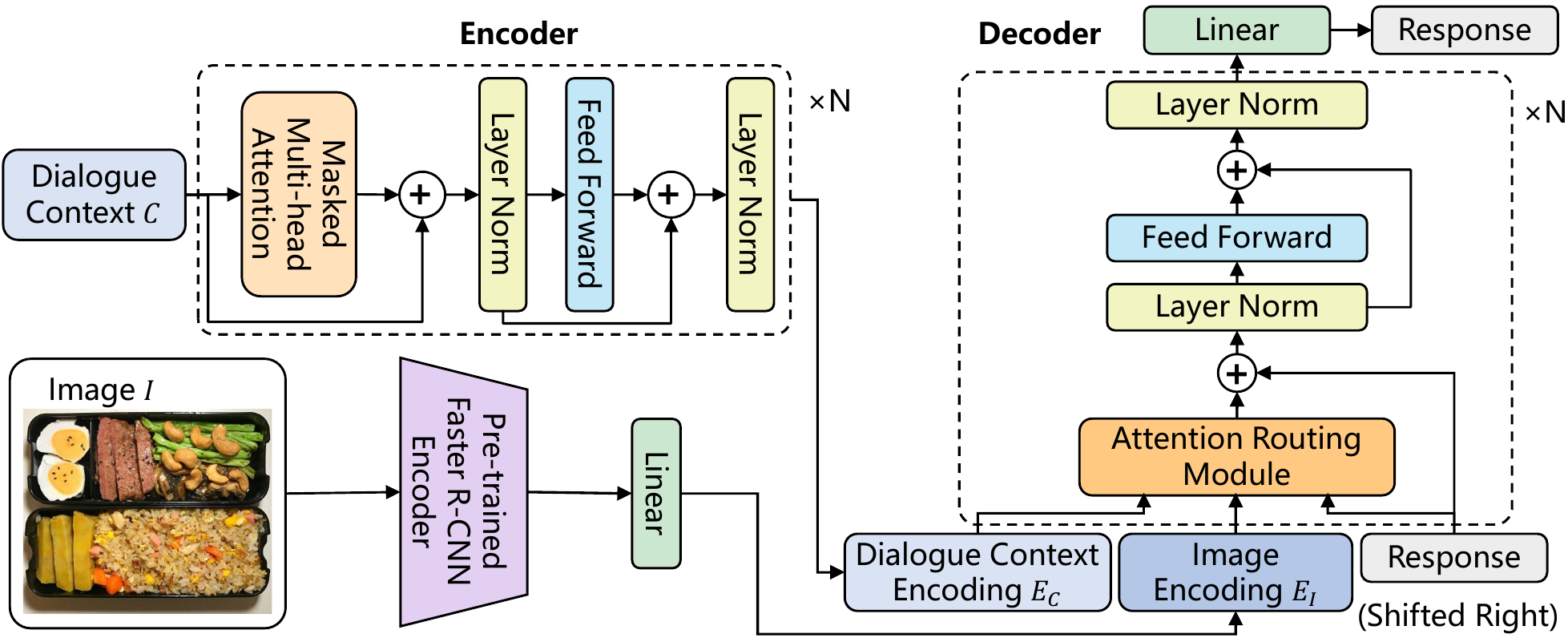}
    \caption{Overview of multi-modal dialogue generation model.}
    \label{fig:model}
\end{figure*}

In \textsc{MMChat-hf}, we only retain dialogue sessions that are annotated as ``true'' in all the above three aspects. This results in 19.90K dialogue sessions as well as 52.66K images. More statistics are shown in Table~\ref{tab:mmachat_hf_info}. In \textsc{MMChat-hf}, dialogues contains 4.07 utterances on average and utterances contains 11.93 characters on average, which are longer than dialogues in \textsc{MMChat} (see Table~\ref{tab:data_info}). All these annotations will be released under our data release protocol.

\subsection{Data Analysis}

We try to reveal the semantics of images contained in \textsc{MMChat}. Specifically, a Faster R-CNN model \cite{Ren2015Faster_rcnn,lu2019vilbert} trained with attribute head on the Visual Genome \cite{krishnavisualgenome} dataset is used to detect objects in each image \footnote{We use the pre-trained model provided by \url{https://github.com/peteanderson80/bottom-up-attention}}. The regions where any class detection confidence exceeds a threshold (0.2 in our case following \cite{Anderson2017up-down}) are selected to further detect the specific object labels. We follow the work of \newcite{Anderson2017up-down} to use an object and attribute vocabulary with the size of 1600 and 400, respectively. An average count of 11.42 objects is detected in each image. This indicates that images in our dataset contain rich semantic information and thus are informative. An example dialogue session, together with its associated images, is shown in Table~\ref{tab:ex2}.

%% file: sec/model.tex
\section{Dialogue Generation on \textsc{MMChat}} \label{sec:model}

This section starts by formally defining the task of MMDS (Section~\ref{sec:task}). Moreover, we propose to use the attention routing mechanism~\cite{zheng2020pre} to capture the issue of sparsity (Section~\ref{sec:model_detail}).

\begin{table*}[t!]
\centering
\resizebox{\textwidth}{!}{%
\begin{tabular}{lccccccc}
\toprule
Model & BLEU-2 & BLEU-3 & BLEU-4 & Dist-1 & Dist-2 & Ent-1 & Ent-2 \\
\midrule
Seq2Seq         & 2.830 & 1.376 & 0.805 & 2.63 & 33.92 & 6.00 & 9.47  \\
Seq2Seq+PIMG  & 2.928 & 1.469 & 0.888 & 2.73 & 34.34 & 6.01 & 9.45  \\
                & (+3.46\%) & (+6.76\%) & (+10.31\%) & (+3.80\%) & (+1.24\%) & (+0.17\%) & (-0.21\%) \\
Seq2Seq+IMG   & 3.001 & 1.588 & 1.006 & 2.82 & 35.38 & 6.07 & 9.52 \\
                & (+6.04\%) & (+15.41\%) & (+24.97\%) & (+7.22\%) & (+4.30\%) & (+1.17\%) & (+0.53\%) \\

% Seq2Seq + IMG (1st Turn)   & 7.071 & 2.751 & 1.367 & 0.815 & 4.22 & 41.27 & 6.08 & 9.34 \\
% Seq2Seq + IMG (multi Turn) & 8.492 & 3.366 & 1.894 & 1.264 & 5.35 & 45.08 & 5.99 & 9.14 \\
\midrule
Human Reference  & N/A & N/A & N/A & 9.09 & 48.77 & 6.69 & 9.64 \\
\bottomrule
\end{tabular}}
\caption{Evaluation Results on \textsc{MMChat}. Relative improvements compared to the Seq2Seq baseline is shown in parentheses.}
\label{tab:results}
\end{table*}

\begin{table*}[t!]
\centering
%\resizebox{\textwidth}{!}{%
\begin{tabular}{lccccccc}
\toprule
Model &  BLEU-2 & BLEU-3 & BLEU-4 & Dist-1 & Dist-2 & Ent-1 & Ent-2 \\
\midrule
Seq2Seq         & 3.779 & 2.405 & 1.641  & 5.35 & 45.62 & 6.11 & 9.26  \\
Seq2Seq+PIMG  & 4.576 & 3.094 & 2.230 & 5.04 & 42.61 & 6.01 & 9.14  \\
                & (+21.09\%) & (+28.65\%) & (+35.89\%) & (-5.79\%) & (-6.60\%) & (-1.64\%) & (-0.21\%) \\
Seq2Seq+IMG   & 4.818 & 3.381 & 2.541 & 5.75 & 45.35 & 6.05 & 9.15 \\
                & (+27.49\%) & (+40.58\%) & (+54.84\%) & (+7.48\%) & (-0.59\%) & (-0.98\%) & (+1.19\%) \\

% Seq2Seq + IMG (1st Turn)   & 7.071 & 2.751 & 1.367 & 0.815 & 4.22 & 41.27 & 6.08 & 9.34 \\
% Seq2Seq + IMG (multi Turn) & 8.492 & 3.366 & 1.894 & 1.264 & 5.35 & 45.08 & 5.99 & 9.14 \\
\midrule
Human Reference  & N/A & N/A & N/A & 6.17 & 47.43 & 5.98 & 9.00 \\
\bottomrule
\end{tabular}%}
\caption{Evaluation Results on \textsc{MMChat-hf}. Relative improvements compared to the Seq2Seq baseline is shown in parentheses.}
\label{tab:results_hf}
\end{table*}

\subsection{Task Definition} \label{sec:task}

The task of MMDS is to learn a function $f$ that can map textual contexts $\mathcal{C}$ (e.g., dialogue histories) and multi-modal contexts $\mathcal{I}$ (e.g., images, audio or video) into dialogue responses $Y$, i.e., learn $f:\{\mathcal{C, I}\} \mapsto Y$. In this study, we focus on the image modality in $\mathcal{I}$, i.e., $\mathcal{I}$ is composed of a set of images $\{I_n\}_{n=1}^N$.

\subsection{Dialogue Generation Model} \label{sec:model_detail}

The Seq2Seq architecture is used as our backbone to build a multi-modal dialogue generation model. As shown in Figure \ref{fig:model}, two encoders are used to respectively encode the textual context $\mathcal{C}$ and image context $\mathcal{I}$ into encoded representations $E_{\mathcal{C}}$ and $E_{\mathcal{I}}$. An attention routing module is utilized to merge $E_{\mathcal{C}}$ and $E_{\mathcal{I}}$ in the decoder, and the response $Y$ is decoded auto-regressively.

\subsubsection{Encoder}
The encoder for the textual context $\mathcal{C}$ is parameterized with the Transformer architecture \cite{vaswani2017attention} (12 layers, 12 attention heads, and 768 hidden states). To improve the generation quality, we initialize its weights using a pre-trained GPT model \cite{radford2018improving}. Utterances in the dialogue history are concatenated using a special token ``[SEP]'', and $E_{\mathcal{C}}$ is obtained by feeding the concatenated token sequence into the textual encoder.

The encoder for the image context $\mathcal{I}$ is implemented as the Faster R-CNN model with ResNet-101 backbone. The weights of this encoder are pre-trained on the Visual Genome dataset and fixed in the training process \footnote{We have released the pre-trained weights}. Specifically, a feature vector with the size of 2048 is extracted from each image region. The top-50 high confidence regions are used to produce $E_{\mathcal{I}}$ with a linear layer to adjust the feature-length, i.e., the resulting $E_{\mathcal{I}}$ contains 50 features, each has a length of 768.

\subsubsection{Decoder}

We implement the dialogue decoder with the Transformer architecture and share its weights with our textual encoder. To tackle the sparsity issue, we equip the dialogue decoder with the attention routing mechanism~\cite{zheng2020pre} to balance the contribution of each region feature. Specifically, given the encoding of the dialogue context $E_\mathcal{C}$, image context $E_\mathcal{I}$, and previous decoded tokens $E_{\mathcal{\mbox{pre}}}$, three attention routes are computed as:
\begin{align}
    O_\mathcal{C} & = \mbox{MHA}(E_{\mathcal{\mbox{pre}}}, E_\mathcal{C}, E_\mathcal{C}), \label{eq:text_cxt} \\
    O_\mathcal{I} & = \mbox{MHA}(E_{\mathcal{\mbox{pre}}}, \gamma E_\mathcal{I}, \gamma E_\mathcal{I}), \label{eq:img_cxt} \\
    O_{\mathcal{\mbox{pre}}} & = \mbox{MMHA}(E_{\mathcal{\mbox{pre}}}, E_{\mathcal{\mbox{pre}}}, E_{\mathcal{\mbox{pre}}}), \label{eq:pre}
\end{align}
where $\gamma \in [0,1]$ is a hyper-parameter to re-scale $E_\mathcal{I}$. MHA and MMHA represent masked and unmasked multi-head attention, respectively, in which $E_{\mathcal{\mbox{pre}}}$ serves as the query. The results of each attention operation are averaged before proceeding to the next sub-module: 
\begin{equation}
    O_{\mbox{merge}} = \frac{O_\mathcal{C} + O_\mathcal{I} + O_{\mathcal{\mbox{pre}}}}{3}. % + E_{\mathcal{\mbox{pre}}}.
\end{equation}
Note that the attention route on image features (i.e., Eq. \ref{eq:img_cxt}) assigns different weights to different image regions. This facilitates more flexible control over image features in the decoding process and thus helps ease the sparsity issue.

%% file: sec/experiment.tex
\begin{CJK}{UTF8}{gbsn}
\begin{table*}[tbp]
    \small
    \centering
    \begin{tabular}{cl}
    \toprule
    Input Image & Dialogue Contexts and Generated Responses \\ \midrule
    \begin{minipage}{.3\textwidth} % 1656
      \includegraphics[scale=0.4]{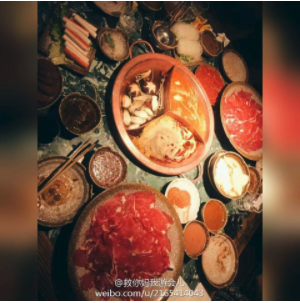}
    \end{minipage} & \makecell[l]{\textbf{Context}: ``在拜托啦学长那里看到你，欢迎来我们福建玩，\\ 坐标福建漳州。不过现在在外上学'' \\
    (``I saw you on the channel "Please! Seniors.". Welcome to Fujian! \\ I am from Zhangzhou, Fujian, but studying out of Fujian.'') \\
    \textbf{Human}: 厦门。\\ (Xiamen.) \\ \textbf{Seq2Seq}: 啊，那还真的好啊，我已经回来了。\\ (Aha! That's great! I have been back.) \\ \textbf{Seq2Seq+PIMG}: 我是漳州的学妹。\\ (I am your younger female schoolmate from Zhangzhou.) \\ \textbf{Seq2Seq+IMG}: 原来是这样，下次有机再来\\福建找你玩哈。\\ (I see. I'll look you up, if I ever have a chance to visit Fujian again.) } \\  \midrule
    \begin{minipage}{.3\textwidth} % 1285
      \includegraphics[scale=0.24]{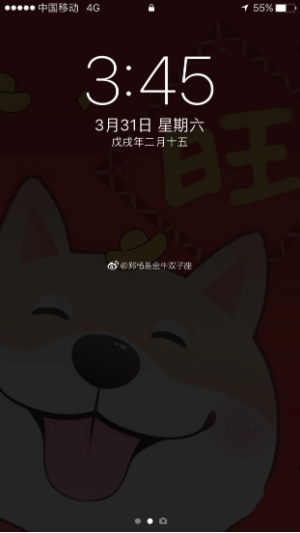}
    \end{minipage} & \makecell[l]{\textbf{Context}: ``你又睡那么晚'', ``美甲！'' ``你有空了就来撒'' \\ (``Why you go to sleep so late again?'', ``Just back from a nail salon.'', \\ ``Shall we meet if you have spare time?'') \\ \textbf{Human}: 好。\\ (Okey.) \\ \textbf{Seq2Seq}: 我就在四小时以左右开始睡不着觉！其他时候都是两三点。\\ (I found it is hard to fall asleep at 4 o'clock this morning. \\ This used to be 2 or 3 o'clock.) \\ \textbf{Seq2Seq+PIMG}: 上完课就去上，没时间了。\\ (I'll go there after class. There is no time.) \\ \textbf{Seq2Seq+IMG}: 好滴！国庆之后就能见到了。\\ (Okey. I suppose to meet you soon after the national day.)} \\
    \bottomrule
    \end{tabular}
    \caption{Example outputs (with English translations) generated by our multi-modal dialogue systems.}
    \label{tab:ex3}
\end{table*}
\end{CJK}

\section{Experiments}

Experiments are performed to assess both our model and datasets. Specifically, we train our model on both \textsc{MMChat} and \textsc{MMChat-hf}. For \textsc{MMChat}, we sample 4.0K and 2.0K dialogue sessions for testing and validation, respectively, and for \textsc{MMChat-hf}, we sample 1.0K and 1.0K dialogue sessions for testing and validation, respectively. Two baselines are also implemented in our study.

\subsection{Implementation Details}

In our proposed dialogue model (referred to as \textbf{Seq2Seq+IMG}), the encoder and decoder are 12-layer transformers with 768-dimensional hidden states and 12 attention heads. For the position-wise feed-forward networks, 3,072-dimensional inner states are used. The Adam optimizer is used to train our model with $\beta_1=0.9$, $\beta_2=0.98$ and $\epsilon=10^{-9}$. The maximum learning rate is set to 1.0e-4. The training starts with a warmup step of 1,000, and the learning rate is annealed proportionally to the inverse square root of the step number. The batch size is set to 360, and the training iterators 60 epochs. A character-level vocabulary of size 13,084 is used. Other settings of our Transformer model follow the work of \newcite{radford2018improving}.

We share the weights of the encoder and decoder in the dialogue model and initialize these weights using a pre-trained GPT model \cite{wang2020chinese}. The pre-training corpus contains about 0.5 billion tokens, and the pre-training process lasts for a week on 8 GTX1080Ti GPUs. We use the top-K ($K=20$) decoding scheme with beam search for the dialogue model in the inference phase. The beam size and length penalty are 4 and 2.0, respectively. The training of our model lasted for about 24 hours on 2 NVIDIA Tesla V100 GPUs. The number of parameters for our model (textual encoder and decoder, not including the image encoder) is 191.01M. Note that due to the large computation loads to utilize the pre-training based method in our model and our baselines, we inherit most of the hyper-parameter settings from the previous studies, such as \newcite{radford2018improving}, and skip the hyper-parameter tuning process. Moreover, for fair comparisons, we use a fixed set of hyper-parameters in all our experiments (including all the baselines).

\subsection{Baselines}

We also implement two baselines to validate our dataset and model: \textbf{Seq2Seq}: a transformer-based Seq2Seq model is built with only textual inputs; \textbf{Seq2Seq+PIMG}: an image-grounded dialogue model is built with a single pooled image representation. Specifically, an max-pooling operation is applied to $E_{\mathcal{I}}$, and the pooled vector is added to each representation vector in $E_{\mathcal{C}}$. The attention route to $E_{\mathcal{I}}$ (i.e., Eq. \ref{eq:img_cxt}) is not applied. Note that the first baseline does not use image contexts, and the second baseline does not model the sparsity phenomenon.

For fair comparisons, all baselines employ the same architecture, hyper-parameter setting and initialization scheme with our model \textbf{Seq2Seq+IMG}. 

\subsection{Metrics}

We use the following metrics: \textbf{BLEU} \cite{papineni2002bleu} measures the n-gram (n=2,3,4) overlaps between generated and reference responses; \emph{Distinct} (\textbf{Dist}) \cite{li2016diversity} measures the proportion of unique n-gram in the generated response (n=1,2);  \emph{Entropy} (\textbf{Ent}) \cite{zhang2018generating} measures how evenly the empirical n-gram (n=1,2) distribution is:
\begin{equation}
    \mbox{Ent} = \frac{1}{\sum_w F(w)} \sum_{w \in V} F(w) \log \frac{F(w)}{\sum_w F(w)},
\end{equation}
where $V$ is the set of all n-grams and $F(w)$ is the frequency of n-gram $w$. Note that both distinct and the entropy measure the diversity of the generated responses.

\subsection{Results}

Table~\ref{tab:results} and Table~\ref{tab:results_hf} shows the results on \textsc{MMChat} and \textsc{MMChat-hf}, respectively. Table~\ref{tab:ex3} lists some example outputs of our dialogue generation models and baselines. Generally speaking, our Seq2Seq+IMG outperforms both baselines on most metrics. The exception is on the entropy metric: the difference between Seq2Seq and Seq2Seq+IMG on both datasets is marginal (approximately 1\%). 

Based on the above results, we can observe that: 1) Incorporating image contexts in dialogue models helps to produce better responses. Specifically, our model obtains 24.97\% and 54.84\% relative improvements on the BLEU-4 score compared to the text-only baseline Seq2Seq on \textsc{MMChat} and \textsc{MMChat-hf}, respectively. Meanwhile, similar improvements are also identified when comparing Seq2Seq+PIMG and Seq2Seq. These results validate our motivation to incorporate multi-modal features in the dialogue generation model and prove that \textsc{MMChat} (as well as \textsc{MMChat-hf}) can be used to build image-grounded dialogue models. 2) Our model, Seq2Seq+IMG, obtains greater relative improvement on BLEU than Seq2Seq+PIMG. This indicates that explicitly modeling the sparsity phenomenon helps to further improve the dialogue generation performance, and \textsc{MMChat}/\textsc{MMChat-hf} facilitates the study of such a phenomenon.

Moreover, by comparing results on \textsc{MMChat} and \textsc{MMChat-hf}, we find that: 1) Models trained on \textsc{MMChat-hf} generally receive higher BLEU and distinct scores than those trained on \textsc{MMChat-hf}; 2) Incorporating image information on \textsc{MMChat-hf} introduce higher improvement on the BLEU score comparing to the improvement observed on \textsc{MMChat} (e.g., BLEU-4 is improved 54.84\% on \textsc{MMChat-hf} while improved 24.97\% on \textsc{MMChat}). These results indicate that filtering out low-quality images and dialogues that are irrelative to their associated images (see Section~\ref{sec:annotation}) do help build a better dataset.

Note that the diversity improvement of our model Seq2Seq+IMG is not significant compared to the baseline Seq2Seq, particularly on the \textsc{MMChat-hf} dataset. This may be because the generated responses are bounded by more contexts (i.e., images). \footnote{Also note that models receive higher distinct scores on \textsc{MMChat-hf} than \textsc{MMChat}, which could, to a large extent, because the test set of \textsc{MMChat-hf} is smaller than that of \textsc{MMChat}.}.